\documentclass[]{bytedance_seed}

\usepackage[toc,page,header]{appendix}
\usepackage{tablefootnote}
\usepackage{amsmath}
\usepackage{amsfonts}
\usepackage{amssymb}
\usepackage{graphicx}
\usepackage{subcaption}
\usepackage{booktabs} % For better table rules
\usepackage{array}    % For column formatting
\usepackage{pdflscape} % For landscape orientation
\usepackage{cleveref}
\usepackage{pdfpages}

\crefname{table}{Table}{Tables} % Sets the singular and plural forms

\usepackage{minitoc}
\usepackage{colortbl}
\usepackage{xcolor}
\usepackage{amssymb}
\usepackage{tcolorbox}
\tcbuselibrary{skins, breakable, listings}

\definecolor{keywordcolor}{rgb}{0.7, 0.1, 0.1}   % red
\definecolor{tacticcolor}{rgb}{0.0, 0.1, 0.6}    % blue
\definecolor{commentcolor}{rgb}{0.4, 0.4, 0.4}   % grey
\definecolor{symbolcolor}{rgb}{0.0, 0.1, 0.6}    % blue
\definecolor{sortcolor}{rgb}{0.1, 0.5, 0.1}      % green
\definecolor{attributecolor}{rgb}{0.7, 0.1, 0.1} % red

\title{Seed-Prover 1.5: Mastering Undergraduate-Level Theorem Proving via Learning from Experience}

\author{ByteDance Seed AI4Math}

\abstract{
Large language models have recently made significant progress to generate rigorous mathematical proofs. In contrast, utilizing LLMs for theorem proving in formal languages (such as Lean) remains challenging and computationally expensive, particularly when addressing problems at the undergraduate level and beyond. In this work, we present \textbf{Seed-Prover 1.5}, a formal theorem-proving model trained via large-scale agentic reinforcement learning, alongside an efficient test-time scaling (TTS) workflow. Through extensive interactions with Lean and other tools, the model continuously accumulates experience during the RL process, substantially enhancing the capability and efficiency of formal theorem proving. Furthermore, leveraging recent advancements in natural language proving, our TTS workflow efficiently bridges the gap between natural and formal languages. Compared to state-of-the-art methods, Seed-Prover 1.5 achieves superior performance with a smaller compute budget. It solves \textbf{88\% of PutnamBench} (undergraduate-level), \textbf{80\% of Fate-H} (graduate-level), and \textbf{33\% of Fate-X} (PhD-level) problems. Notably, using our system, we solved \textbf{11 out of 12 problems} from Putnam 2025 within 9 hours. Our findings suggest that scaling learning from experience, driven by high-quality formal feedback, holds immense potential for the future of formal mathematical reasoning.
}
\checkdata[Project Page]{\url{https://github.com/ByteDance-Seed/Seed-Prover}}
\checkdata[Model ID]{Doubao-Seedprover-1.5}

\lstdefinestyle{mypromptstyle}{
    language={},
    basicstyle=\ttfamily\footnotesize\color{black},
    keywordstyle=\color{black},
    identifierstyle=\color{black},
    stringstyle=\color{black},
    commentstyle=\color{black},
    breaklines=true,
    numbers=none,
    columns=fullflexible, % 使用 fullflexible 通常比 flexible 复制效果更好
    keepspaces=true,
    escapechar=|,
    showstringspaces=false,
    upquote=true,
    extendedchars=false % 建议先设为 false，防止 utf8 字符（如中文或符号）导致编译错误
}

\newtcblisting{finalpromptbox}[2][]{
    listing engine=listings,
    enhanced,
    breakable,
    colback=white,
    colframe=black!75,
    boxrule=0.8pt,
    arc=2pt,
    title={\textbf{#2}},
    coltitle=white,
    fonttitle=\sffamily,
    attach boxed title to top left={xshift=3mm, yshift*=-\tcboxedtitleheight/2},
    boxed title style={colback=black!75},
    top=12pt, bottom=12pt, left=8pt, right=8pt,
    listing only,
    listing style=mypromptstyle, % 调用上面定义的样式
    #1
}

\begin{document}
\maketitle
\vspace{-20pt}
%不需要目录就注释掉 注意目录不要和第一页放在一块 要有\newpage
%\newpage
%\tableofcontents
%\newpage

\begin{figure}[h]
	\centering
        \includegraphics[width=0.9\linewidth]{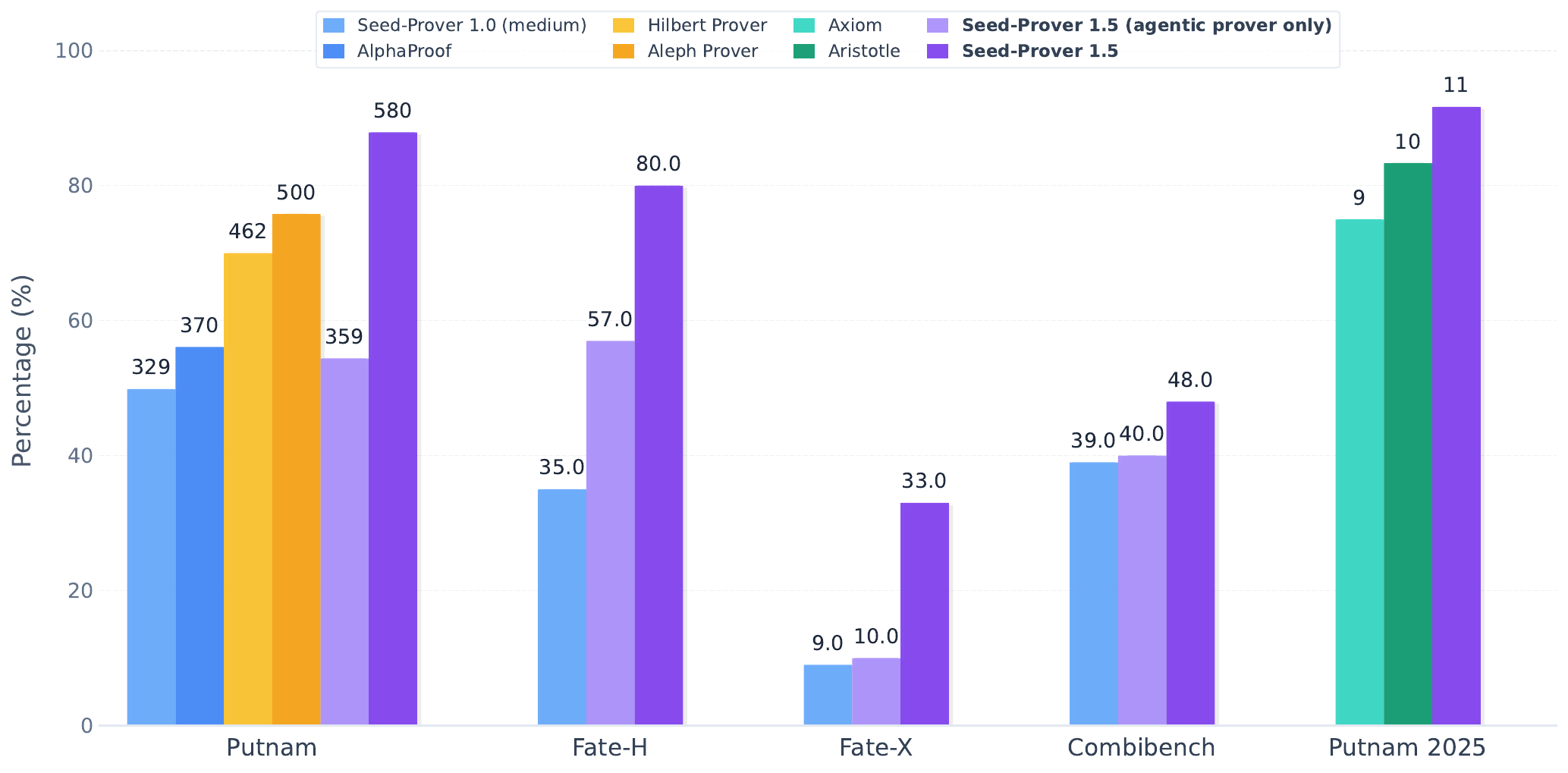}
        \vspace{-.3cm}
		\caption{Performance of Seed-Prover 1.5 compare to other state-of-the-art provers.}
		\label{fig:agent}
\end{figure}

\section{Introduction}

Leveraging Lean\citep{lean4} for mathematical theorem proving offers fully trustworthy verification, effectively eliminating hallucinations and logical errors pervasive in natural language proof \citep{ying2024internlmmathopenmathlarge}. Consequently, formal verification is widely regarded as a potential paradigm shift for mathematical research. However, recent progress in natural language reasoning by LLMs challenges the necessity of this shift. Through specialized RL training and inference workflows, LLMs can not only derive correct answers using natural language, but also ensure a high degree of correctness and rigorousness in the proving process \citep{huang2025winninggoldimo2025,deepseek-math-v2,gao2025longhorizonreasoningagentolympiadlevel}. In stark contrast, a massive gap in capability and efficiency persists between formal and natural language proving. For example, while DeepSeek-Math-V2 \citep{deepseek-math-v2} achieved near-perfect scores on the Putnam 2024 competition, AlphaProof \citep{alphaproof} solved only 56\% of the full Putnam benchmark which is on average simpler than the 2024 set, despite consuming massive computational resources (approximately 500 TPU-days per problem). 
If LLMs can achieve high rigor in natural language proof, while formal proof continues to impose a heavy performance tax, one might question: Is pursuing formal theorem proving with LLMs still a viable and valuable path?

We argue that the answer is yes. In addition to the potential of formal theorem proving on AI4Math,
Lean provides a unique advantage: it serves as a fully verifiable environment where models can freely explore, accumulate experience, and undergo large-scale \textbf{Agentic RL} training with ground-truth feedback. Distinct from the step-level mode (one interaction per tactic)\citep{alphaproof,Aristotle,xin2025bfsproverscalablebestfirsttree, xin2025scaling} or the whole-proof mode (long thinking followed by a single interaction)\citep{kimina,ren2025deepseekproverv2advancingformalmathematical,chen2025seed}, an agentic prover equipped with
experiential learning dynamically adjust its interaction granularity and master auxiliary tools. Therefore,
it represents a superior paradigm in terms of both capability and efficiency. Yet, training such an agentic
prover via large-scale RL remains an underexplored frontier area. In this work, we demonstrate the
scaling potential of this approach.

Furthermore, given the significant improvement in the ability of LLMs to generate rigorous natural language proofs, it is natural to leverage this capability to assist and accelerate formal proving. We train a sketch model to generate lemma-based Lean sketches, establishing an effective bridge between natural language and formal language. The sketch model acts as a hierarchical problem decomposer, enabling the agentic prover to solve sub-problems in parallel, which underpins our design of an efficient test-time scaling workflow.

In this work, our contributions are as follows:
\begin{enumerate}
    \item We established an environment integrating Lean with various tools. Through large-scale RL, we trained an agentic prover to learn optimal interaction strategies and tool usage for formal theorem proving.
    \item We trained a sketch model using Rubric RL to bridge natural language proofs with formalization. Based on this, we implemented a highly efficient test-time workflow.
    \item We present Seed-Prover 1.5, a system that pushes the boundaries of automated formal proving while maintain a moderate compute budget. It achieves state-of-the-art performance across key benchmarks: solving 88\% of Putnam, 80\% of Fate-H, and 33\% of Fate-X problems. Notably, our system successfully proved 11 out of 12 problems from the 2025 Putnam Competition within a 9-hour window. These performances show LLM-based formal proving has narrowed the gap with natural language reasoning in terms of both capability and efficiency at the undergraduate and graduate levels.

\end{enumerate}

\section{Related Works}
Automated theorem proving is a challenging task in artificial intelligence \citep{wu2022autoformalization,polu2022formal,zhengminif2f}. Several systems have integrated Lean 4 with large language models to achieve IMO-level mathematical proving capabilities \citep{alphaproof,chen2025seed,Aristotle}. Broadly, large language model-based provers fall into two categories: step-level interaction\citep{wu2024internlm25stepproveradvancingautomatedtheorem,xin2025bfsproverscalablebestfirsttree,xin2025scaling,alphaproof,Aristotle} and whole-proof generation\citep{kimina,xin2025deepseekproverv,ji2025leanabellproverv2verifierintegratedreasoningformal,ren2025deepseekproverv2advancingformalmathematical,lin2025goedel,lin2025goedelproverv2scalingformaltheorem,deltaprover,chen2025seed,stepfunprover2025}. Step-level models generate a single tactic for a given proof state and interact with Lean incrementally at each tactic step, while whole-proof models produce complete Lean code and interact with Lean only once via the full code. However, both paradigms suffer from inefficient interaction with Lean: step provers interact too frequently, while whole-proof models interact too sparsely.
In contrast to prior works, Seed Prover 1.5 is an agent-based prover that interacts with Lean through lemmas, offering a more balanced and efficient interaction.
Recently, Hilbert \citep{varambally2025hilbert} employs a powerful general reasoning model for informal mathematical proving and a specialized Lean model for formal verification, achieving strong performance on PutnamBench \citep{tsoukalasputnambench}. In contrast to Hilbert, our approach utilizes rubric reinforcement learning \citep{gunjal2025rubricsrewardsreinforcementlearning} to train a sketch model that bridges natural language proofs and Lean sketches, while training an extremely strong agent-based Lean provers.
\section{Approach}

\subsection{Agentic Prover}

\begin{figure}[h]
	\centering
        \includegraphics[width=1\linewidth]{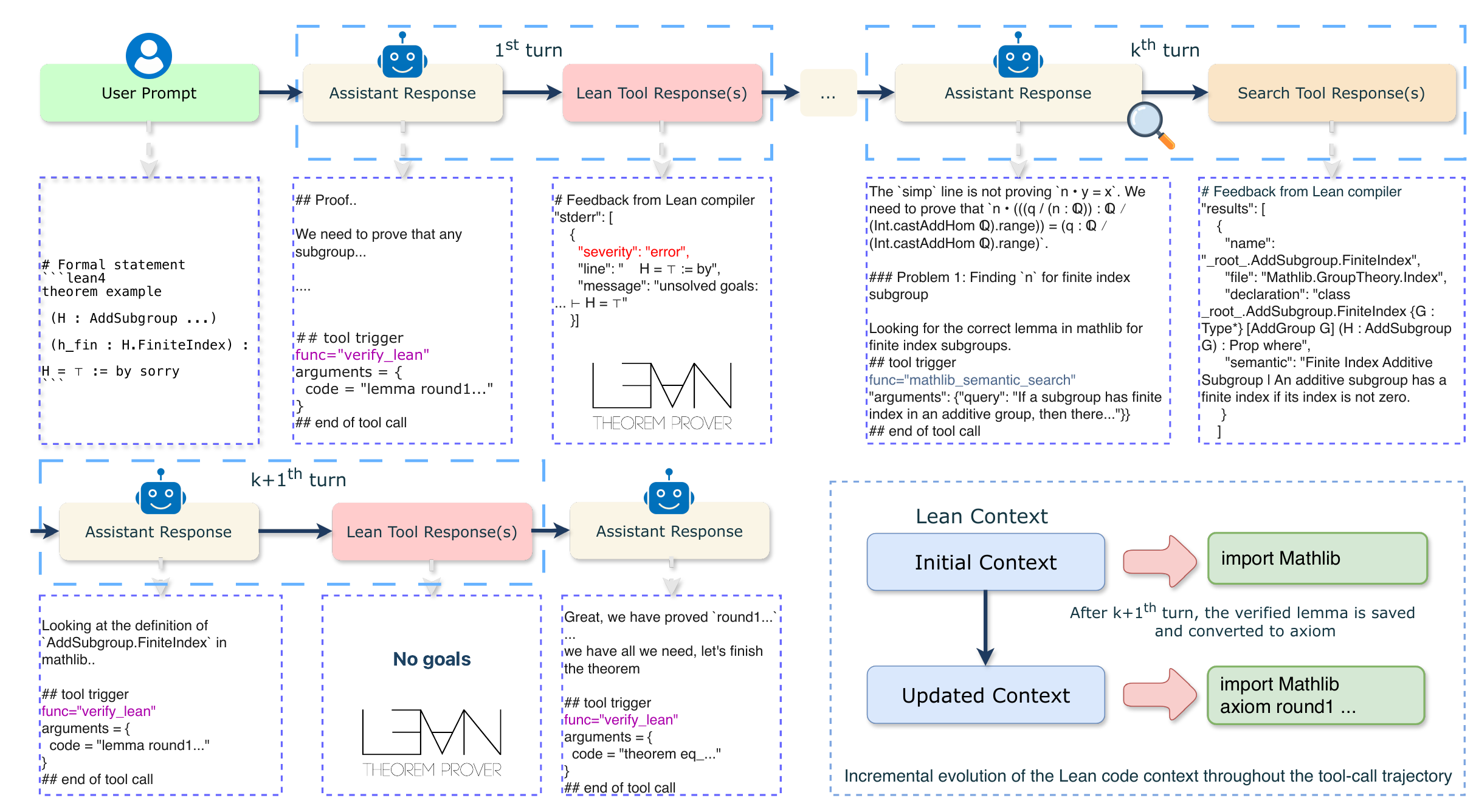}
		\caption{An example of Seed-Prover 1.5 agentic prover running on FATE-H problem.}
		\label{fig:agent}
\end{figure}

Unlike the prior agentic prover~\cite{stepfunprover2025}, which generates an entire proof and repeatedly evaluates it using the Lean compiler, we propose a more efficient strategy that incrementally invokes multiple tools to construct formal proofs step by step. Figure~\ref{fig:agent} illustrates the inference workflow of our agentic prover.
Once a lemma is successfully compiled, it is cached in memory and reused in subsequent reasoning steps, eliminating the need to regenerate previously verified code. 
This incremental caching mechanism enables more efficient utilization of the context window compared to approaches based on whole-proof generation.
This representation offers several advantages:
\begin{itemize}
    \item \textbf{Alignment with Modular Proofs}: It seamlessly integrates with our previously proposed lemma-style proof representation~\cite{chen2025seed}.
    \item \textbf{Decomposition of Complexity}: It relieves the model of the necessity to generate the whole proof. Instead, the model can focus on resolving the immediate sub-goal.
    \item \textbf{Context Efficiency}: By sequentially caching valid lemmas, we significantly reduce context overhead compared to approaches that must iteratively regenerate the full proof history.
    \item \textbf{Flexible Inference Control}: It enables the implementation of diverse inference strategies, such as pruning irrelevant intermediate steps or backtracking to restart the conversation at specific points.

\end{itemize}
% By using this tool-use prover alone, we have already outperform our previous Seed Prover 1.0 Medium prover
% We have trained a tool-use prover that 
% To address these challenges, we have trained a state-of-the-art model to effectively leverage Lean verification together with auxiliary tools, resulting in efficient proving workflow.
% \textcolor{red}{also talk about performance by a single model: Our model is able to achieve... within.xxx.sampling..}

\paragraph{Tools}
Our tools can be grouped into three categories: \textsc{Lean} verification, Mathlib search\footnote{\url{https://github.com/leanprover-community/mathlib4}}, and Python execution.
For \textsc{Lean} verification, we employ LooKeng~\cite{chen2025seed}, a REPL\textsuperscript{4}-based Python interface that compiles Lean proofs and returns structured feedback to the model.
We permit the model input a lemma at each time instead of a whole proof. The statement header and proved lemmas are stored in the running context.
For Mathlib search, we use embedding-based retrieval to identify relevant theorems by semantic similarity, calibrated to a fixed Mathlib commit (i.e., v$4.22.0$) to ensure consistency and reproducibility. 
The search tools can return the most relevant \textit{theorem}, \textit{lemma}, or \textit{def} declarations whose semantics align closely with the given query.
Finally, we provide a Python execution interface that allows the model to generate and run Python scripts, enabling numerical experiments and other computational checks within the proving trajectory.

\paragraph{Inference}
The input prompt is a Lean formal statement with an optional natural-language proof or other auxiliary instructions.
The model begins by reasoning in natural language and calling tools to validate its intermediate proof steps.
Multiple tool calls may occur within a single turn, and we impose no restrictions on the number or ordering of such calls.
As illustrated in Figure~\ref{fig:agent}, the model may invoke ``\textit{Mathlib search}''  to explore available theorems and understand how they can be applied.
Once sufficient context is gathered, the model proceeds to construct a formal proof of the lemma for Lean verification.
Because the formal goal is known in advance, generation terminates either when the final theorem is successfully verified or when the interaction budget (maximum number of turns or maximum sequence length) is exhausted. For all subsequent experiments, we configure our tool-use agent with a maximum sequence length of 64K and a limit of 28 tool calls.
This process is viewed as Pass@1 in our setting.
We can also apply light inference introduced in Seed-Prover \citep{chen2025seed} by applying self-summarizing when the interaction budget is exhausted.

\subsection{Post-training of Agentic Prover}

\paragraph{Cold Start}

We build upon our prior model in Seed-Prover 1.0~\cite{chen2025seed}, and further post-train it to function as an agentic tool-use model tailored in \textsc{Lean} environment.
To support this transition, we construct in-house synthetic training data and use it to perform supervised fine-tuning (SFT), enabling the model to learn our tool invocation patterns and interactions specific to the proving environment.

\paragraph{RL training}
Following our prior work~\cite{chen2025seed}, we design several task formats for RL training, including proving directly from the formal statement, proving conditioned on a natural-language proof sketch, and proving based on a summary of previous failed attempts.
Our training set comprises a mixture of of a combination of publicly available datasets~\cite{albalak2025bigmathlargescalehighqualitymath,numina_math_datasets,peng2025criticlean,ying2024leanworkbooklargescalelean,kimina} and in-house formalized math textbooks including Graduate Texts in Mathematics. 
To construct a high-quality RL dataset, we further filter examples by evaluating the SFT model under a \text{light inference} setting (Pass@$4\times 8$). 
In this setting, if the model fails to complete a proof within a trajectory, it performs self-summarization over this trajectory and initiates a new trajectory conditioned on the summary.
We exclude any example that the model successfully proves more than three times, thereby focusing RL training on sufficiently challenging instances where additional learning signals are most beneficial.
We also remove samples that cannot be proved by the SFT model under any prompting strategy.
Notably, if a formal statement is provable when conditioned on a summarization prompt but not under the direct proving prompt, we retain such examples under the direct proving prompt, as this sample is provable and potentially could improve the model's efficiency.
We implement our RL algorithm based on VAPO~\cite{vapo} and follow similar approach in ReTool~\cite{feng2025retool} to enable tool-integrated reinforcement learning. 
We adopt a simple outcome-based reward function: the model receives a reward of $1$ if a valid proof is completed and verified by the \textsc{Lean} compiler, and $-1$ otherwise. 
\begin{equation}
    \mathcal{L}_{\text{PPO}}(\theta) = -\frac{1}{\sum_{i=1}^G |o_i|} \sum_{i=1}^G \sum_{t=1}^{|o_i|} \min \left( r_{i,t}(\theta)\hat{A}_{i,t}, \text{clip}\left(r_{i,t}(\theta), 1 - \textcolor{black}{\varepsilon_{\text{low}}}, 1 + \textcolor{black}{\varepsilon_{\text{high}}}\right) \hat{A}_{i,t} \right),
    \label{vapo}
\end{equation}
where $G$ is the training batch size, $o_i$ is the trajectory of the $i^{th}$ sample,  $\hat{A}_{\cdot, t}$ is the estimated advantage at time step $t$, and $\varepsilon$ is a
hyperparameter for the clipping range. 
$r_{\cdot,t}(\theta) = \frac{\pi_\theta(a_{t} | s_{t};~\mathcal{T})}{\pi_{\theta_{\text{old}}}(a_{t} | s_{t};~\mathcal{T})}$ is the probability ratio and $\pi_\theta$ represents the rollouts with interleaved tool calls and tool responses $\mathcal{T}$.
During training, the model will interact with the environment to execute tool calls and obtain the tool response for multi-turn generation. 
Our model demonstrates substantial improvement after RL training (Table \ref{tab:efficient_eval}), with the single model significantly outperforming the previous medium workflow.
We expect that iteratively leveraging the RL-trained model to collect additional data may further improve performance, which we leave for future investigation.

\subsection{Sketch Model}

% Large Language Models primarily acquire knowledge and reasoning patterns within the natural language (NL) space. This grants them inherent flexibility and efficiency compared to formal languages, albeit at the expense of strict rigor. While LLMs achieve state-of-the-art performance in informal mathematical reasoning, generating complete Lean proofs for Putnam-level problems in a single pass remains intractable.

To harness the strong natural language proving capabilities, we propose to train a sketch model. This model synthesizes a lemma-style Lean sketch\citep{chen2025seed} from a formal statement with its natural language proof. By generating auxiliary lemmas (initially admitted via \texttt{sorry}) and organizing them into a main proof body, the model decomposes the proposition into N independent sub-goals. Since Lean guaranties the high-level structural soundness, evaluation shifts to the quality of the lemmas: a sketch is valuable if its lemmas are mathematically valid, and the most challenging lemma is strictly easier than the original proposition.

To train this sketch model, we utilize VAPO~\cite{vapo} with a hybrid reward signal. The Lean compiler ensures the correctness of the sketch structure, while an LLM-as-a-Judge Rubric acts as a semantic value model, employing Long Chain-of-Thought (Long-CoT) to achieve better generalization than scalar-based models.
We require the natural language prover to verify each lemma, immediately rejecting the sketch (i.e. natural language quality score is $-1$) if any lemma is mathematically invalid. 
We find using the natural language prover to disprove is cheaper than using a theorem prover.
Subsequently, we prompt an LLM to evaluate the overall sketch quality, considering factors including alignment with the NL proof, decomposition granularity, difficulty reduction, and Lean junk value analysis. 
Finally, we fuse these metrics into a strict binary reward ($+1/-1$):  Let $N_{\text{lemmas}}$ be the number of generated lemmas, $S_{\text{FL}}$ be the lean verification score, and $S_{\text{NL}}$ be the natural language quality score. To encourage sufficient decomposition and high quality, we define the reward function $R$ as follows:

\begin{equation} \tag{2}
    R = 
    \begin{cases} 
      1 & \text{if } N_{\text{lemmas}} \ge 3 \land S_{\text{FL}} \ge 0 \land S_{\text{NL}} \ge 0.7, \\
      -1 & \text{otherwise.}
    \end{cases}
\end{equation}
Then the sketch model is optimized using VAPO similar to Equation~\ref{vapo}.
The prompt we used for lemma verification and rubric reward and the sample output of the sketch model is shown in Appendix~\ref{appendix:sketch}.

\subsection{Test-Time Workflow}

Proving complex mathematical theorems in Lean typically entails thousands of lines of code, rendering monolithic, single-pass generation computationally intractable. To address this, we implement a hierarchical test-time workflow that facilitates multi-agent collaboration. This system orchestrates context management, problem decomposition, and sub-goal assignment to tackle complex proofs effectively.

Seed-Prover 1.5 orchestrates three specialized agents:
\begin{enumerate}
    \item \textbf{Natural Language Prover}: An LLM optimized for natural language proving (initialized from Doubao-Seed-1.6). Its role is to generate rigorous, lemma-style natural language proofs to guide the formalization.
    \item \textbf{Sketch Model}: A translation agent trained to convert natural language proofs into lemma-Style Lean sketches, effectively bridging the gap between informal reasoning and formal syntax.
    \item \textbf{Agentic Lean Prover}: The tool-integrated agent (described in the previous section) responsible for verifying every individual lemma.
\end{enumerate}

Given a Lean statement, the Natural Language Prover first generates a proof in natural language, which the Sketch Model then converts into a lemma-style Lean sketch. For each unsolved lemma within the sketch, the Agentic Prover attempts to prove or disprove it, under a compute budget of Pass@3 × 3.
If a proof cannot be found, the system recursively performs the "natural language proof → Lean sketch" decomposition. If a lemma is disproved, the system reverts to the Sketch Model to refine the sketch. This process repeats until every leaf node (sub-lemma) in the search tree is successfully proved by the Lean Prover, or the maximum search depth is reached.

\section{Experiments}

We evaluated our agentic prover and test-time workflow using the following benchmarks under Lean v4.22.0:

\begin{itemize}
    \item PutnamBench \citep{tsoukalasputnambench}: Consists of 660 problems from the William Lowell Putnam Mathematical Competition (from 1962 to 2024). It evaluates the model's ability to solve undergraduate-level mathematical problems.
    \item FATE \citep{jiang2025fate}: FATE-H contains 100 problems at the level of honors course exams or graduate-level difficulty. FATE-X contains 100 problems at the level of PhD qualifying exams or beyond. These benchmarks evaluate the model's capacity to solve mathematical problems with graduate level math knowledge and beyond.
    \item CombiBench \citep{liu2025combibench}: CombiBench is a benchmark specifically centered on combinatorial problems, where the problems often involve newly-defined concepts. We use it to assess the model's capabilities in combinatorics, which is often a shortcoming for LLMs and formal provers. However, we discovered significant formalization issues within this dataset. We list the performance here for reference.
    \item IMO \& Putnam 2025: We use these two competitions to measure our model's performance. These two competitions have no chances of data leakage. 
    \item Erdős\footnote{\url{https://www.erdosproblems.com/}}: To test if our model is capable of proving frontier math conjectures, we use problems from Erdős problem sets. We collected a subset of Erdős problems from FormalConjecures Project\footnote{\url{https://github.com/google-deepmind/formal-conjectures/tree/main/FormalConjectures}} and our in-house expert labeling, removing those with formalization errors (e.g. Erdős-74, 590, 591).
\end{itemize}

\begin{figure}[t!]
    \centering
    \begin{subfigure}{0.48\textwidth}
        \centering
        \includegraphics[width=\linewidth]{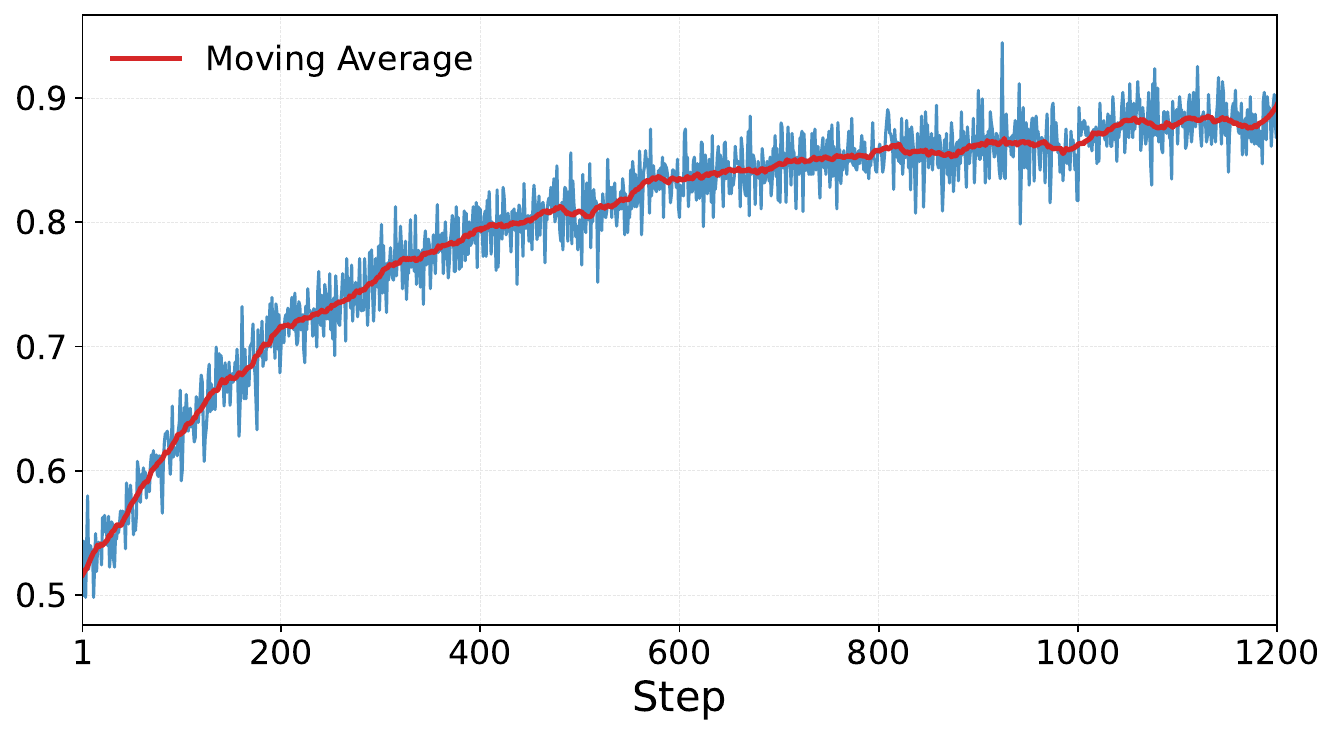}
        \caption{Batch-level training accuracy.}
        \label{fig:training_acc}
    \end{subfigure}
    \hfill
    \begin{subfigure}{0.48\textwidth}
        \centering
        \includegraphics[width=\linewidth]{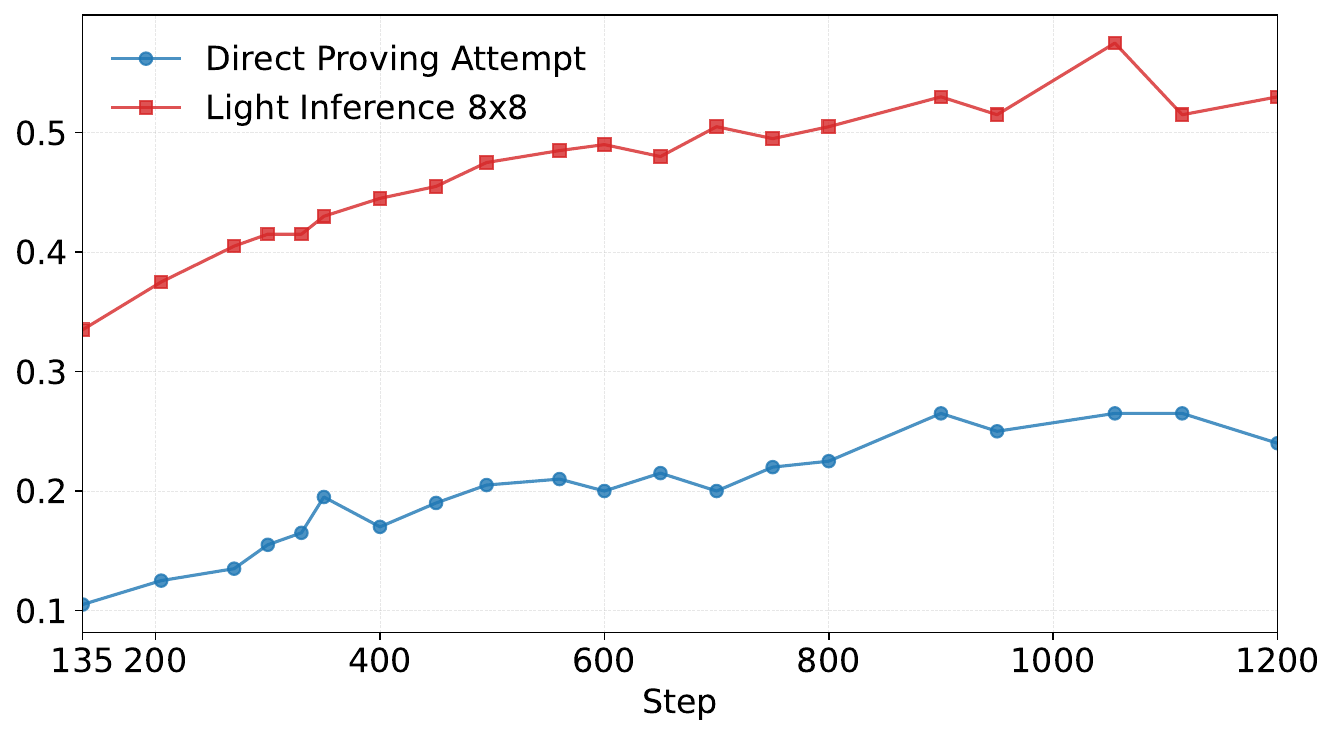}
        \caption{Accuracy on the test set Putnam-200.}
        \label{fig:putnam200}
    \end{subfigure}
    \caption{Training dynamics and evaluation metrics over 1200 RL training steps.}
    \label{fig:main}
\end{figure}

\subsection{Scaling Behavior of the Agentic Prover Training}
We monitor the performance of our agentic prover during large-scale RL training.

\begin{figure}[h]
    \centering
    % ---- Row 1 ----
    \begin{subfigure}{0.48\textwidth}
        \centering
        \includegraphics[width=\linewidth]{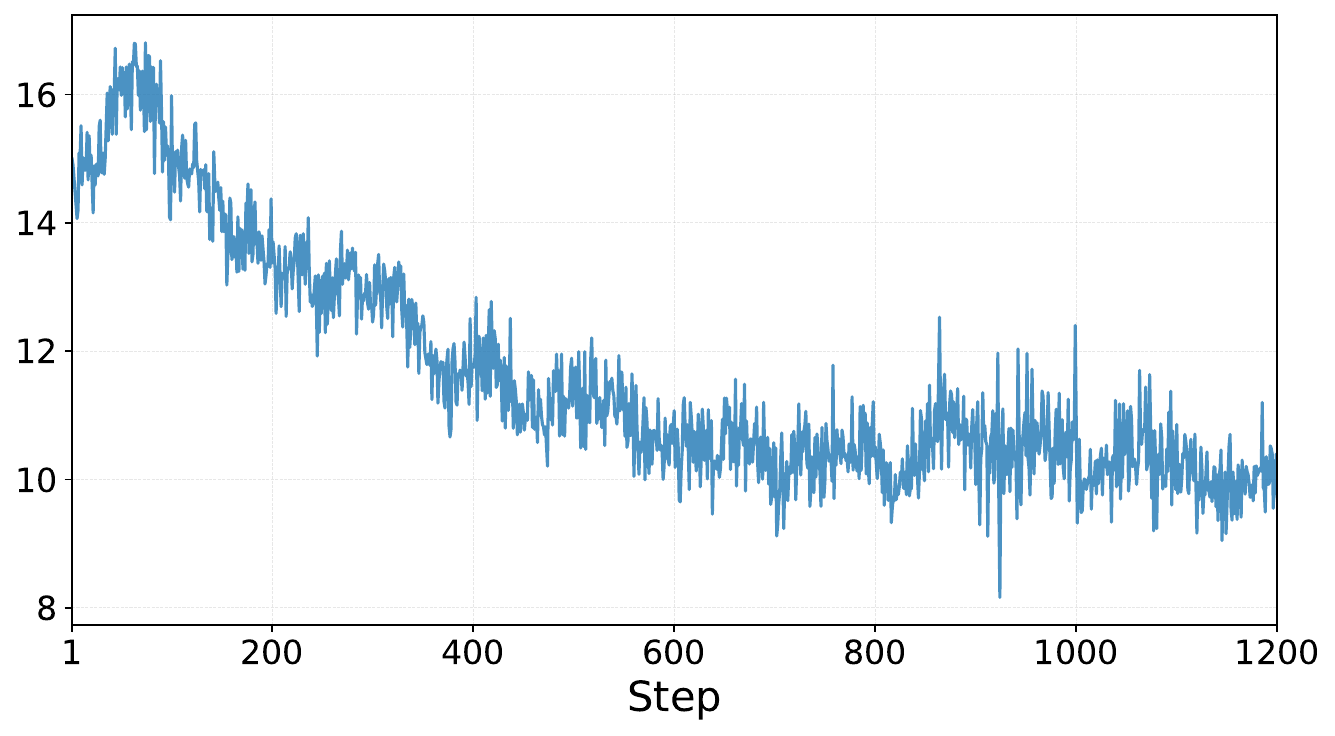}
        \caption{Average number of function calls}
        \label{fig:4a}
    \end{subfigure}
    \hfill
    \begin{subfigure}{0.48\textwidth}
        \centering
        \includegraphics[width=\linewidth]{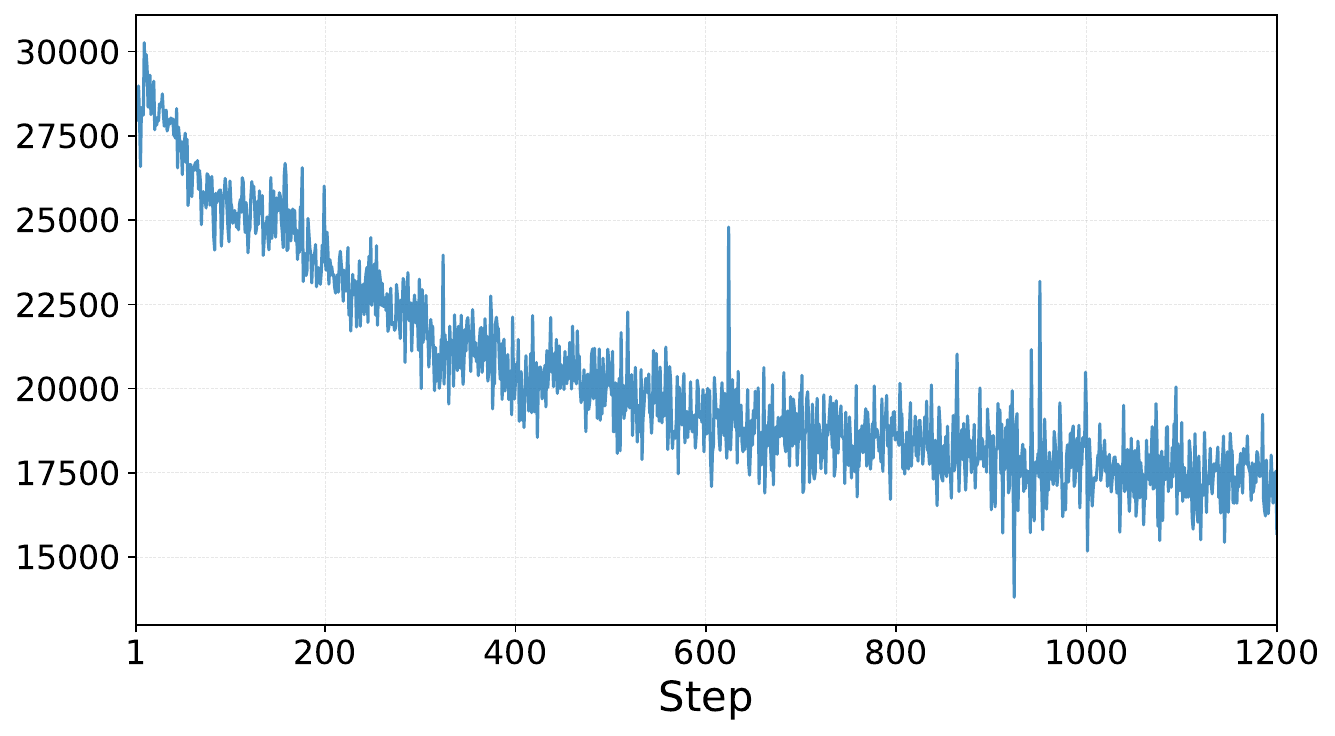}
        \caption{Average total sequence length}
        \label{fig:4b}
    \end{subfigure}

    % \medskip  % vertical space between rows

    % ---- Row 2 ----
    \begin{subfigure}{0.48\textwidth}
        \centering
        \includegraphics[width=\linewidth]{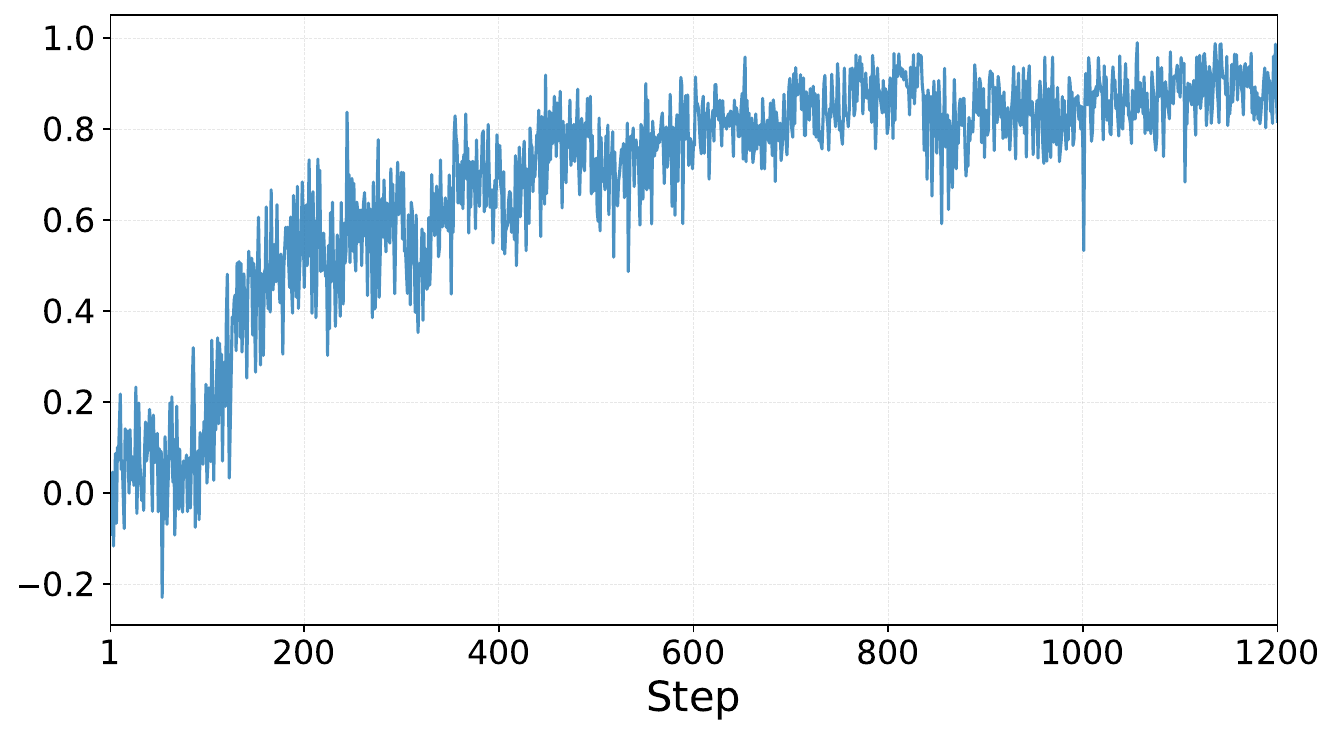}
        \caption{Score of sample responses with lengths 16K to 32K}
        \label{fig:4c}
    \end{subfigure}
    \hfill
    \begin{subfigure}{0.48\textwidth}
        \centering
        \includegraphics[width=\linewidth]{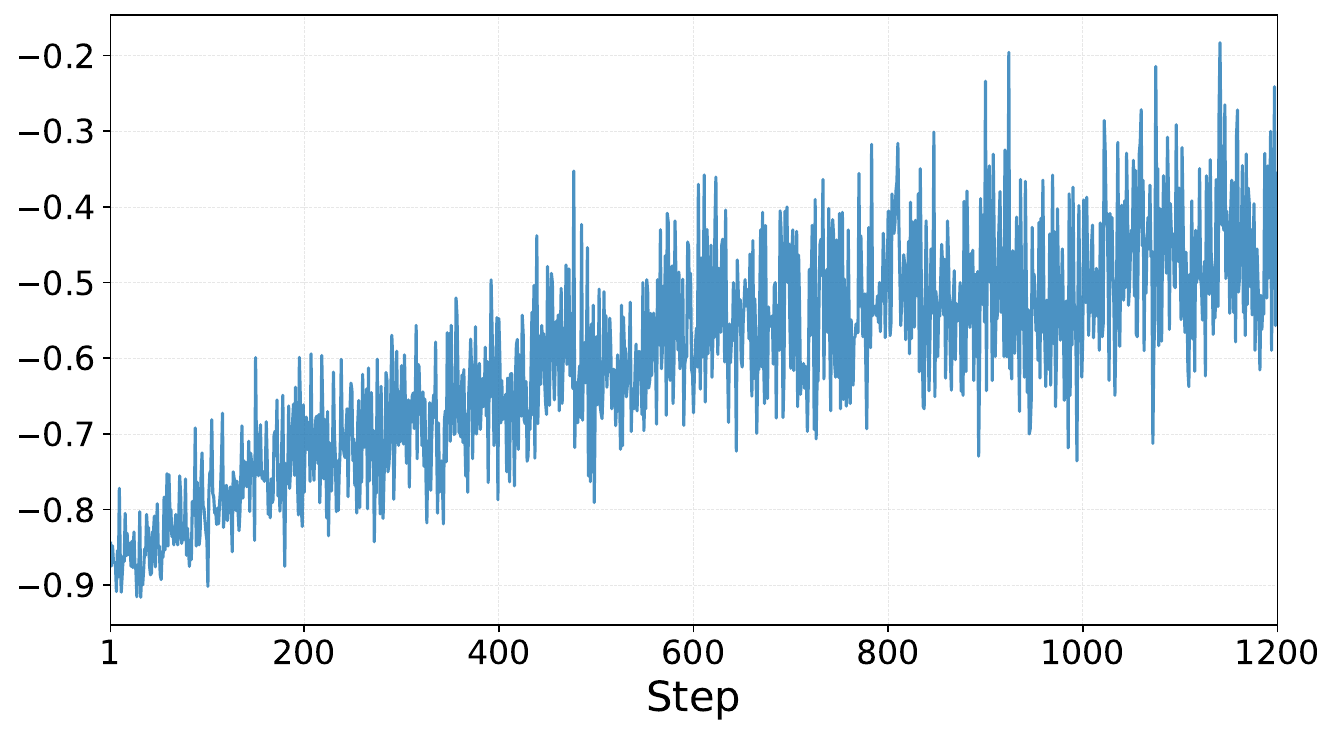}
        \caption{Score of sample responses with lengths 32K to 64K}
        \label{fig:4d}
    \end{subfigure}

    \caption{Other training metrics in our agentic RL training.}
    \label{fig:other_metrics}
\end{figure}

\paragraph{RL Training Dynamics}
As shown in Figure \ref{fig:training_acc}, RL training accuracy increases from approximately 50\% at initialization to nearly 90\% at more than 1000 steps. 
We believe this improvement is enabled by two key factors: (i) a curated training dataset that aligns closely with the proof tasks, and (ii) the accurate reward signal from Lean verification. 
Consequently, a reward approaching 90\% suggests that reinforcement learning effectively helps the model extract maximal benefit from the training data. Figure \ref{fig:other_metrics} provides a deeper view into the agent's behavioral change during RL training. 
Figure \ref{fig:4a} and \ref{fig:4b} reveal a significant optimization in efficiency: the average number of function calls drops from approximately 15 to 10, which correlates with a consistent reduction in average sequence length (from $\sim$28k to $\sim$17k tokens, see Figure \ref{fig:4b}). 
This suggests the model is learning to use tools more strategically, avoiding redundant or ``trial-and-error'' invocations.
Despite the above reduction, the model's reasoning capability improves. 
Figure \ref{fig:4c} and \ref{fig:4d} track the scoring metrics for samples with longer response lengths (16K–64K).
The improvement over optimization steps indicates that the agent is also increasingly capable of managing complex, long-horizon problems.
However, the persistence of negative scores within the 32K–64K range suggests that effectively reasoning over extremely long contexts remains a challenge.

\begin{figure}[t!]
    \centering
    \begin{subfigure}{0.48\textwidth}
        \centering
        \includegraphics[width=\linewidth]{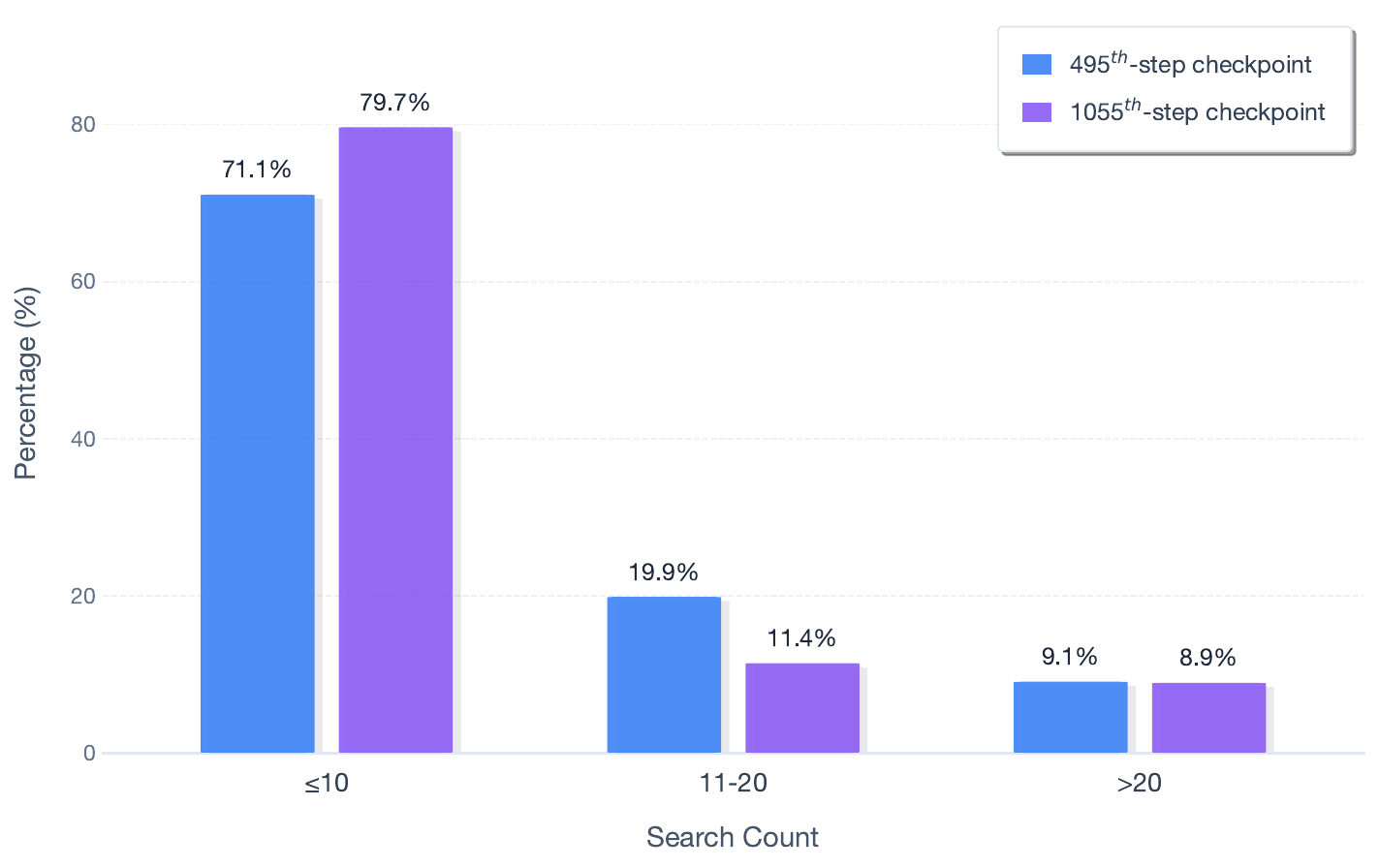}
        \caption{Putnam}
        \label{fig:putnam_search_distribution}
    \end{subfigure}
    \hfill
    \begin{subfigure}{0.48\textwidth}
        \centering
        \includegraphics[width=\linewidth]{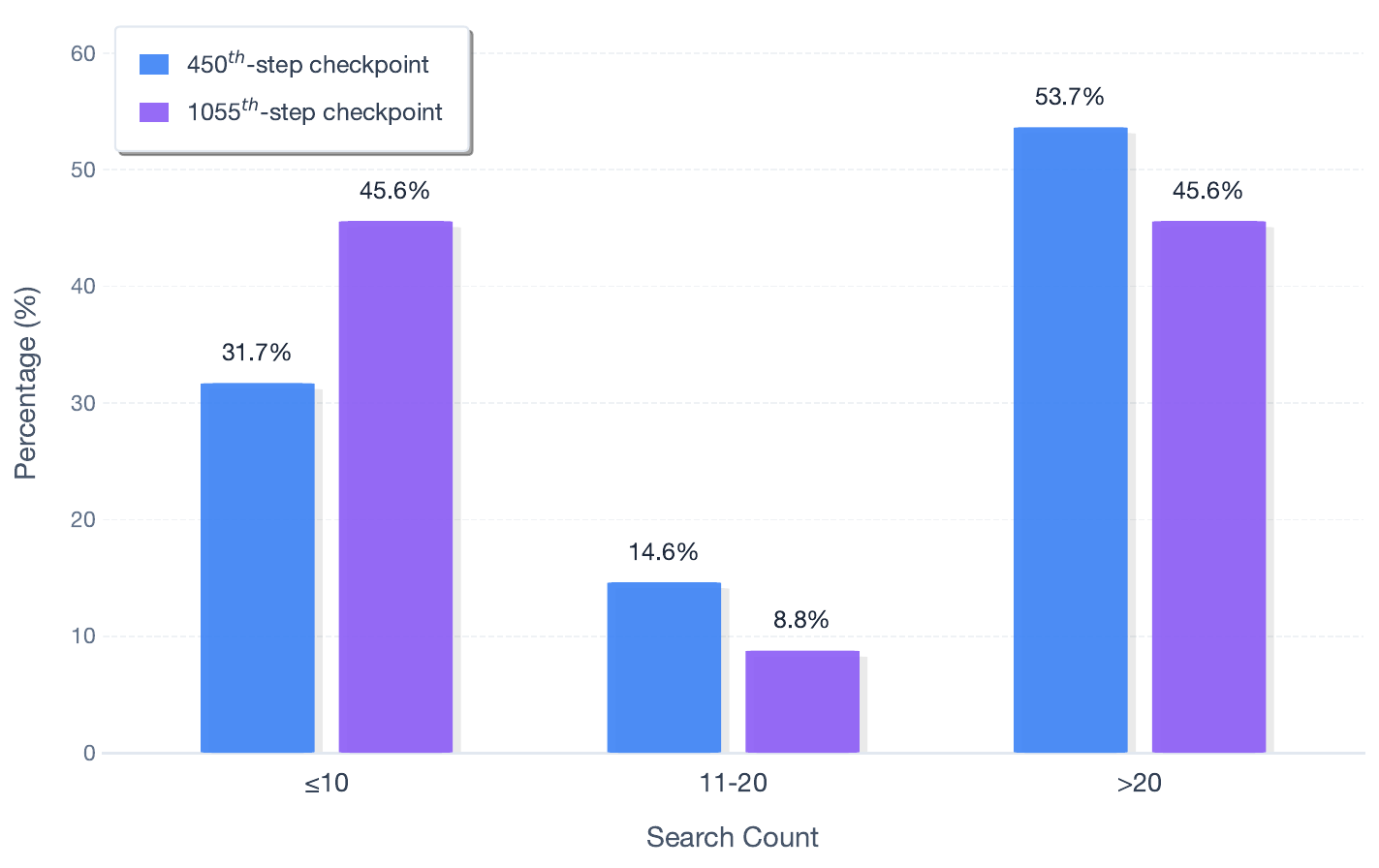}
        \caption{Fate-H}
        \label{fig:fate_search_distribution}
    \end{subfigure}
    \caption{Distribution of search calls for proved samples using different checkpoints. The charts illustrate the percentage of successfully proved samples stratified by the number of search calls ($\le 10$, $11-20$, and $>20$) for the (a) Putnam and (b) Fate-H benchmarks. The number of search tool calls per sample here include all trajectories (i.e., summary-based trajectory)}
    \label{fig:search_change}
\end{figure}

\paragraph{Adaptive Search Behavior}
During RL training, the average number of search tool calls per trajectory remained generally low ($<3$). 
However, we observed a distinct difference between datasets during inference: on Fate-H, the model averaged approximately $10$ search calls per trajectory, whereas on Putnam, it averaged only $1$--$2$. 
This suggests that the model is capable of adapting its tool-call strategy to different scenarios, particularly given that Fate relies heavily on Mathlib search to derive proofs. 
We also observed that the number of search calls decreased in later checkpoints, while performance continued to improve (see Figure \ref{fig:search_change}). 
This trend suggests that RL training not only improves reasoning capabilities but also enables the model to internalize knowledge from search results. 
For example, we found that training cases with significantly decreased response lengths are often able to locate the key lemma faster and identify the correct theorem to complete the proof.

\begin{table}[h!]
\centering
% 1. 增加行高，让表格更舒展
\renewcommand{\arraystretch}{1.4}

\begin{tabular}{lcccc}
\toprule[1.5pt] % 顶部粗线

% 2. 第一行加粗，作为表头
\textbf{Approach}   & \textbf{Budget} & \textbf{Putnam} & \textbf{Fate-H} & \textbf{Fate-X} \\ 

\midrule[1pt] % 中间分割线

% 3. 第一列（Row Header）加粗，数据居中
Goedel-Prover-V2-32B     & pass@$64$  &  86/660        &  2/100           & 0/100              \\
Seed-Prover 1.0 (medium) & 18 H20 days / problem
  &  331/660        &  35/100           & 9/100                 \\ 
\textbf{Seed-Prover 1.5} (agentic prover only) & pass@$8\times8$  &  \textbf{359/660}        &  \textbf{57/100}           & \textbf{10/100}                 \\ 

\bottomrule[1.5pt] % 底部粗线
\end{tabular}

\caption{Performance between Seed-Prover 1.5 agentic prover and Seed-Prover 1.0 medium workflow. }
\label{tab:efficient_eval}
\end{table}

\paragraph{Test-set Performance}
To enable fast development, we select $200$ problems from PutnamBench named Putnam-200 to evaluate different RL model steps.
Figure \ref{fig:putnam200} shows that accuracy on the Putnam-200 subset continues to increase as the training reward increases, for both direct solving (Pass@8x1) and the light inference (Pass@8×8) setting.
This result represents a substantial improvement over our previous configurations, indicating that continued scaling of reinforcement learning leads to sustained gains in both training and test performance. 
As a comparison with Seed-Prover 1.0~\cite{chen2025seed}, we select the best-performing checkpoint at $1055^{\text{th}}$ training step and evaluate its light inference performance on the full PutnamBench and Fate benchmarks, as reported in Table~\ref{tab:efficient_eval}.
All evaluations for the agent prover are conducted with a maximum sequence length of 64K tokens and a maximum of 28 tool calls.
Under a compute budget of Pass@$8 \times 8$ the proposed agent significantly outperforms Seed-Prover 1.0 using the medium workflow, despite the latter consuming substantially more computational resources than light inference. 
Note that the reported results do not leverage longer sequence lengths or additional inference-time strategies such as error pruning, which enable further performance improvements.

\subsection{Evaluation and Scaling Behavior of the Test-Time Workflow}

\begin{figure}[h]
    \centering
    \begin{subfigure}{0.49\textwidth}
        \centering
        \includegraphics[width=\linewidth]{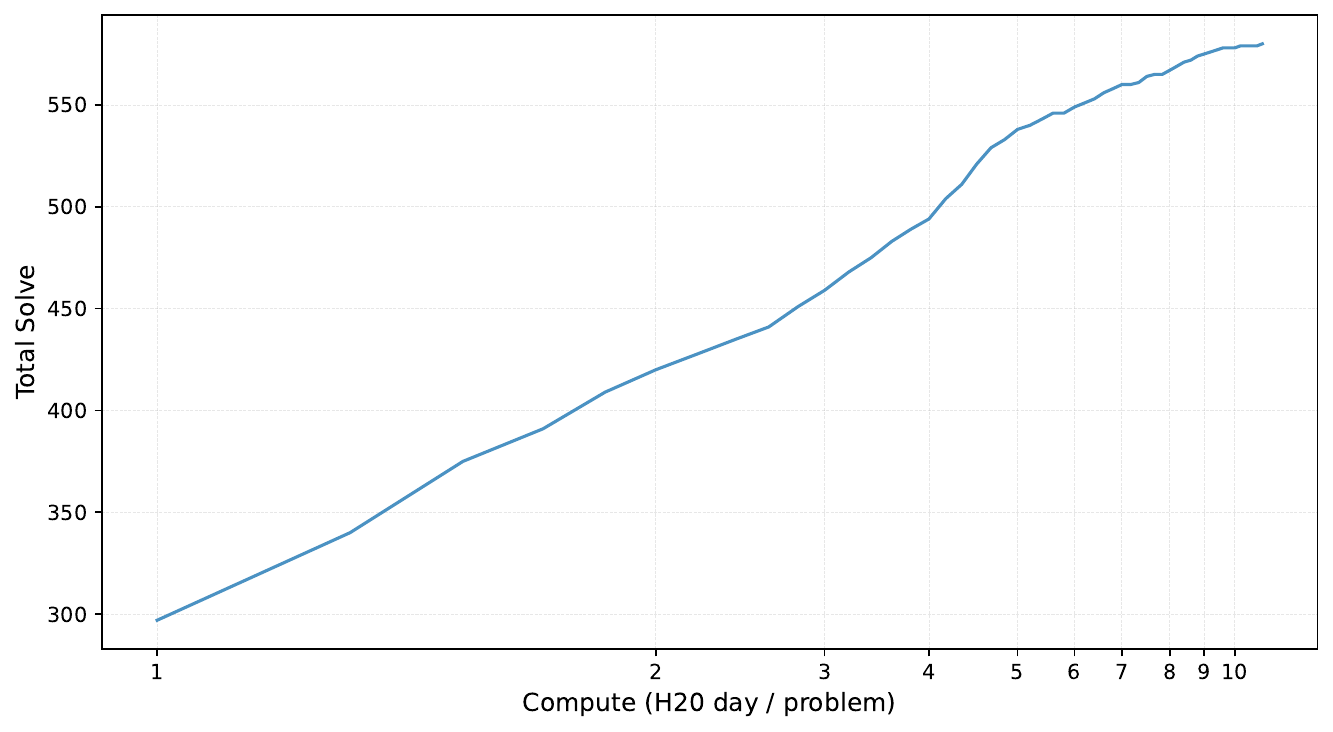}
        \caption{Scaling of solved problems with compute. The number of solved problems on PutnamBench increases log-linearly with respect to the computational budget.}
        \label{fig:putnam_solve_vs_compute}
    \end{subfigure}
    \hfill
    \begin{subfigure}{0.49\textwidth}
        \centering
        \includegraphics[width=\linewidth]{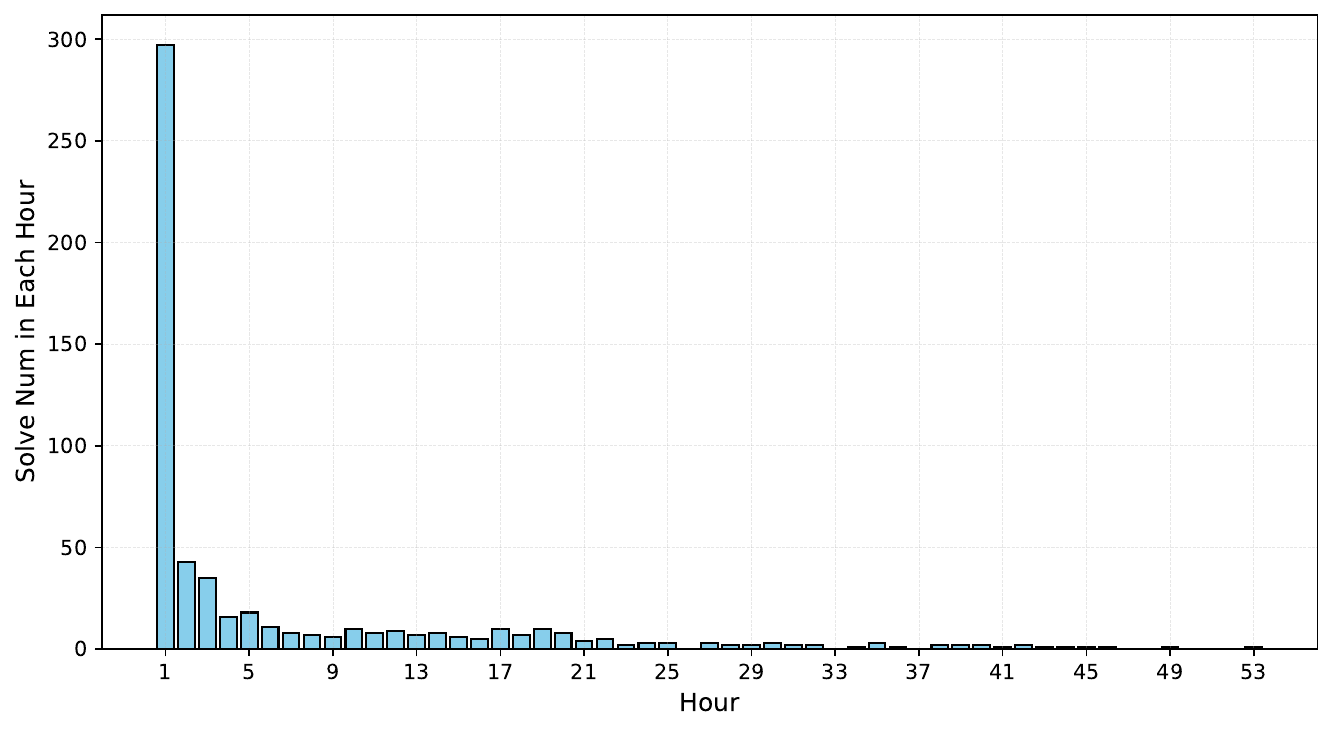}
        \caption{Problem distribution by solution time. The histogram shows the number of problems solved within each specific hour of search.}
        \label{fig:putnam_solve_per_hour}
    \end{subfigure}
    \caption{Test-time Scaling of Seed-Prover 1.5 on PutnamBench.}
    \label{fig:main}
\end{figure}

\paragraph{Scaling} We present the test-time scaling behavior of Seed-Prover 1.5 workflow on PutnamBench. For each problem, we set an initial maximum search depth of 4. If a problem reaches the limit without resolution, we incorporate the lemmas proven during the search into the context and restart the search from scratch. Consequently, this extends the maximum search depth to 8 for each problem. As illustrated in Figure \ref{fig:putnam_solve_vs_compute}, investing more compute (including search width and search depth) leads to a log-linear increase in Seed-Prover 1.5's solve rate. Figure \ref{fig:putnam_solve_per_hour} demonstrates that a significant majority of problems are solved within the first few hours, while a long tail of more challenging problems is discovered as the search duration extends up to the 53rd hour.

\paragraph{PutnamBench \& FATE} Table \ref{tab:seed_prover_results} shows that Seed-Prover 1.5 demonstrates a significant advantage over AlphaProof \citep{alphaproof}, Hilbert Prover \citep{varambally2025hilbert}, and Aleph Prover on PutnamBench. Furthermore, compared to Seed-Prover 1.0, Seed-Prover 1.5 achieves significantly better performance across various benchmarks with a reduced compute budget. Our system solves 87.9\% of problems on PutnamBench and 80\% on Fate-H, demonstrating its proficiency in handling formal proofs for undergraduate and graduate-level mathematics.
However, the system still faces challenges with PhD-level
problems and beyond, primarily due to the increased complexity and limitations in Mathlib support.

\begin{table}[h]
\centering
\renewcommand{\arraystretch}{1.3}

\resizebox{\textwidth}{!}{
\begin{tabular}{llcccc} 
\toprule[1.5pt]
\textbf{Name} & \textbf{Compute Budget} & \textbf{Putnam} & \textbf{Fate-H} & \textbf{Fate-X} & \textbf{Combibench} \\ 
\midrule[1pt] % 中部线

Seed-Prover 1.0 (medium) & 18 H20 days / problem  & 50.4\%         & 35\%           & 9\%            & 39\%           \\ 
AlphaProof               & 500 TPU days / problem & 56.1\%         & -              & -              & -              \\
Hilbert                  & avg pass@1840         & 70.0\%         & -              & -              & -              \\
Aleph Prover             & avg 1834 tool calls   & 75.8\%         & -              & -              & -              \\
\midrule
\textbf{Seed-Prover 1.5}         & 10 H20 days / problem  & \textbf{580/87.9\%}  & \textbf{80\%}  & \textbf{33\%}  & \textbf{48\%}  \\ 

\bottomrule[1.5pt]
\end{tabular}
}
\caption{Performance comparison of Seed-Prover 1.5 against other methods.}
\label{tab:seed_prover_results}
\end{table}

\paragraph{IMO \& Putnam 2025} We evaluated Seed-Prover 1.5 on the IMO 2025 (Table \ref{tab:imo_solve_time}) and Putnam 2025 competition (Table \ref{tab:putnam_solve_time}). In these evaluations, we capped the maximum search depth at 4 while increasing the parallel width to accelerate problem resolution. While Seed-Prover 1.0 required "Heavy" mode to solve 5 out of 6 problems, Seed-Prover 1.5 achieved the same solve rate using compute resources comparable to the 1.0 "Medium" setting (20 H20-days/problem), with a significantly shorter runtime than Seed-Prover 1.0 (Heavy). We also tested on the 12 problems from Putnam 2025; utilizing a maximum compute budget of 40 H20-days/problem, we successfully solved 11 of them within 9 hours\footnote{To be noticed, our prover is not using any `native\_decide' in Putnam, which is unsafe under Lean.}.

\begin{table}[h]
\centering
% 1. 增加行高，让表格更舒展
\renewcommand{\arraystretch}{1.4}

\begin{tabular}{lcccccc}
\toprule[1.5pt] % 顶部粗线

% 2. 第一行加粗，作为表头
\textbf{IMO 2025}   & \textbf{P1} & \textbf{P2} & \textbf{P3} & \textbf{P4} & \textbf{P5} & \textbf{P6} \\ 

\midrule[1pt] % 中间分割线

% 3. 第一列（Row Header）加粗，数据居中
\textbf{Solve Hour} & 16.5   & 0.01     & 5           & 8           & 1           & X           \\ 

\bottomrule[1.5pt] % 底部粗线
\end{tabular}

\caption{Time taken (in hours) for Seed Prover 1.5 to solve IMO 2025 problems. (P2 is solved by
Seed-Geometry)}
\label{tab:imo_solve_time}
\end{table}

\begin{table}[h]
\centering
% 1. 调整行高，保持视觉舒适
\renewcommand{\arraystretch}{1.4}

% 2. 使用 resizebox 自动缩放表格以适应文本宽度 (防止超出页边距)

% 3. 定义列格式：第一列左对齐，其余12列居中 (lcccccccccccc)
\begin{tabular}{l*{12}{c}}
\toprule[1.5pt] % 顶部粗线

% 4. 表头加粗
\textbf{Putnam 2025} & \textbf{A1} & \textbf{A2} & \textbf{A3} & \textbf{A4} & \textbf{A5} & \textbf{A6} & \textbf{B1} & \textbf{B2} & \textbf{B3} & \textbf{B4} & \textbf{B5} & \textbf{B6} \\

\midrule[1pt] % 中间分割线

% 5. 第一列标题加粗
\textbf{Solve Hour}  & 1           & 0.5         & 2           & 4           & X           & 4           & 9           & 6           & 0.5         & 2           & 4           & 3           \\

\bottomrule[1.5pt] % 底部粗线
\end{tabular}

\caption{Time taken (in hours) for Seed Prover 1.5 to solve Putnam 2025 problems.}
\label{tab:putnam_solve_time}
\end{table}

\paragraph{Erdős} Seed-Prover 1.5 solved problem numbers 124, 198, 303, 316, 330, 350, 370, 379, 418, 449, 493, 499, 645, 728 and 958. However, based on our observations, these problems are mathematically relatively simple, or we proved a trivial or simplified version of them due to mis-formalization (these mis-formalized problems are not listed here). Our systems, whether using natural language or formal language, are still some distance away from truly helping to advance research on frontier open mathematical problems, which motivates our pursuit of further research.

\section{Conclusion}
In this paper, we propose Seed-Prover 1.5, a high-performance agentic Lean prover integrates with a sketch model which serves as a bridge between natural language proofs and formal Lean code. Seed-Prover 1.5 shows strong performance on competitive mathematics problems (e.g., Putnam Competition) and graduate-level mathematical tasks (e.g., FATE dataset), thereby laying a solid foundation for proving frontier mathematical problems.
Nevertheless, our system currently cannot make significant mathematical contributions comparable to human experts, and this limitation stems from a critical "dependency issue" inherent to frontier mathematical research. A key distinction lies in the nature of mathematical tasks: IMO or Putnam problems are deliberately designed such that solvers do not require knowledge of specific prior research papers while driving significant progress in mathematics typically hinges on synthesizing insights across a multitude of related papers.
Achieving such progress requires addressing three interconnected challenges: first, identifying the most influential and relevant papers; second, conducting natural language proofs grounded in these works; third, developing scalable approaches to formalizing both the papers themselves and the results derived from them.
Solving these three challenges would enable the large-scale generation of formal mathematical research—an advancement that could ultimately contribute to resolving certain open mathematical conjectures.

\clearpage

\bibliographystyle{plainnat}
\bibliography{main}

\clearpage
\beginappendix
\section{Contributors}
The names are sorted alphabetically. An asterisk * indicates a member who left Seed.

Jiangjie Chen, Wenxiang Chen, Jiacheng Du, Jinyi Hu, Zhicheng Jiang, Allan Jie, Xiaoran Jin, Xing Jin, Chenggang Li, Wenlei Shi, Zhihong Wang, Mingxuan Wang, Chenrui Wei, Shufa Wei, Huajian Xin, Fan Yang*, Weihao Gao, Zheng Yuan, Tianyang Zhan, Zeyu Zheng, Tianxi Zhou, Thomas Hanwen Zhu

\section{Prompts of Rubric RL and Example of Sketch}
\label{appendix:sketch}

\begin{finalpromptbox}{Prompt 1: Atomic Lemma Verification}


**Role and Goal**
* **Role:** You are a rigorous mathematician and an expert Lean 4 formalization engineer.
* **Overall Objective:** Determine if a specific **"Formal Statement"** is **Mathematically Correct** and **Provable**.
* **Key Philosophy:** Treat the "Formal Statement" as an independent proposition. You must verify it using logical reasoning and standard mathematical axioms.

**Context Usage (The Sketch)**
* **Purpose of Sketch:** The provided "Lean Code Sketch" is strictly a **Dictionary for Definitions**.
* **When to use it:** Only refer to the Sketch to resolve the definitions of unknown symbols (e.g., custom `def`, `structure`, `class`, or `instance`).
* **What to IGNORE:** **Ignore all other `lemma` or `theorem` entries inside the Sketch.** They may be hallucinated or incorrect. Do NOT assume they are true, and do NOT cite them in your proof.

**Task Description**

1.  **Symbol Resolution:**
    * Read the **Formal Statement**.
    * If standard math symbols are used (e.g., `Nat`, `Real`, `List`), apply standard mathematical semantics (considering Lean's implementation details).
    * If custom symbols are used, look up their precise definitions in the **Lean Code Sketch**.

2.  **Independent Verification (The Core Task):**
    * **Construct a Proof or Counter-Example:** Attempt to prove the statement from first principles (definitions) or standard mathematical theorems.
    * **Do NOT Circularly Reference:** Never prove the statement by saying "it is already listed in the sketch."
    * **Check for Counter-Examples:** Actively try to break the statement. If a counter-example exists (e.g., "bounded increasing sequence is not necessarily eventually constant"), the statement is **Incorrect**.
    * **Special Check for Infinite Operators:**
        * **Integrals (`∫`):** Check for `IntegrableOn` or `Integrable` hypotheses. If missing, especially on non-compact domains (e.g., `Set.Ici`, `Set.univ`), the integral is likely undefined (junk value). Note: Compact domains with continuous functions are safe.
        * **Infinite Sums (`∑'`):** Check for `Summable` hypotheses. If missing, `∑'` evaluates to `0` (junk value). Verify if this causes a contradiction (e.g., `Positive_LHS ≤ ∑' (divergent) = 0`). If so, the statement is **Incorrect**.

3.  **Junk Value Analysis:**
    * **Consult the Reference:** rigorousy apply the rules from the provided **"Lean 4 Corner Cases and Junk Values"** document.
    * **Check Edge Cases:** Verify if the statement fails due to Lean's total functions (e.g., check `0` denominators, empty lists, infinite sums, or `Nat` subtraction).
    * **Verdict:** If the theorem holds in standard math but fails because Lean returns a specific junk value (e.g., `1/0 = 0` makes the equation false), it is **Incorrect**.

**Assessment Criteria**

* **Correct (Provable):**
    * The statement holds true based on the definitions and standard logic.
    * A valid, step-by-step natural language proof exists.

* **Incorrect (Unprovable):**
    * **Mathematical Falsehood:** A counter-example exists.
    * **Missing Hypothesis:** The statement is not universally true because it lacks a necessary precondition (e.g., missing `Integrable` or `Summable`).
    * **Logic Gap:** The statement cannot be derived from the definitions.
    * **Junk Value Failure:** The statement fails due to Lean's total function handling (e.g., `1/0` or `∑' divergent = 0`).

**Output Format**

Return a single valid JSON object:

{{
  "correctness": "Correct" or "Incorrect",
  "reason": "Concise explanation. If 'Correct', briefly summarize the proof logic. If 'Incorrect', provide the counter-example, the specific logical flaw, or the missing hypothesis (e.g., 'The statement is incorrect because it requires a `Summable` hypothesis, otherwise the sum is 0.').",
  "proof_sketch": "A rigorous Natural Language proof or a concrete counter-example construction. Do not use 'sorry'."
}}

Note: correctness must be "Correct" or "Incorrect".
----------------------------sketch((Definitions)---------------------------------------
{sketch}

----------------------------formal statement to evaluate---------------------------------------
{formal_statement}


--------lean4 doc string maybe helpful for you to understand the theorem--------
{doc_string}


--------docs about Lean 4 Corner Cases and Junk Values  that maybe helpful--------
Lean 4 Corner Cases and Junk Values
This document catalogs important corner cases in Lean 4's mathematical library (Mathlib) where definitions use "junk values" or default values for edge cases. This is not an exhaustive list.
What are Junk Values?
In Lean 4, many mathematical operations are total functions (defined for all inputs) rather than partial functions. When an operation is mathematically undefined or doesn't make sense for certain inputs, Lean assigns a default "junk value" instead of leaving it undefined.
Common Examples
Arithmetic Operations
 * Natural number subtraction: (2 : ℕ) - (3 : ℕ) = 0
   * Natural numbers cannot be negative, so subtraction returns 0 when the result would be negative
   * Notation: ℕ represents natural numbers (0, 1, 2, ...)
 * Division by zero: x / 0 = 0 in many contexts
   * Instead of being undefined, division by zero returns 0
Cardinality
 * Extended cardinality of infinite sets: Nat.encard s = 0 when s is infinite
   * Nat.encard returns the cardinality as a natural number
   * For infinite sets, it returns 0 as a junk value (since infinite cardinality cannot be represented as ℕ)
Real-valued Functions
 * Square root of negative numbers: Real.sqrt x = 0 when x < 0
   * The real square root is only defined for non-negative numbers
   * For negative inputs, it returns 0 as a junk value
 * Logarithm of non-positive numbers: Real.log x = 0 when x ≤ 0
   * The real logarithm is only defined for positive numbers
   * For non-positive inputs, it returns 0 as a junk value
Calculus
 * Derivative of non-differentiable functions: deriv f = 0 when f is not differentiable
   * If a function isn't differentiable at a point, its derivative is defined as 0
Series and Products
 * Divergent infinite sums: ∑' i, f i = 0 when the series doesn't converge
   * The notation ∑' represents an infinite sum (tsum)
   * When a series diverges, it evaluates to 0 by default
   * Note: Summable and HasSum typically mean absolutely summable in Lean, which is a stronger condition than conditional convergence
 * Divergent infinite products: prod' i, f i = 1 when the product doesn't converge
   * The notation prod' represents an infinite product (tprod)
   * When a product diverges, it evaluates to 1 by default
Working with Junk Values
When assessing provability:
 * CRITICAL CHECK: If the Doc String describes a general mathematical fact (e.g., "The derivative of log is 1/x") but the Formal Statement lacks necessary preconditions (e.g., x is not 0 or DifferentiableAt), it is Unprovable.
 * If the statement fails due to a junk value, mark it as Unprovable and explain the specific edge case (e.g., "Fails when x=0 because 0/0=0, not 1").

 Please evaluate the Formal Statement above based on the Sketch and Reference provided. Return the JSON verdict.
\end{finalpromptbox}

\begin{finalpromptbox}{Prompt 2: Proof Strategy Alignment}
### **1. ROLE & GOAL**

You are a Lean 4 proof strategy evaluator. Your primary goal is to assess if an AI-generated "proof sketch" represents a sound and effective proof plan for a given theorem. A good plan decomposes a complex problem into simpler, independent, and solvable sub-problems (lemmas) that facilitate automated theorem proving. You must be concise and follow a strict "fail-fast" workflow.

### **2. GUIDING PRINCIPLE**

Think like a senior mathematician mentoring a student on how to simplify a problem for a computer solver. Is the student's plan (the sketch) logical and does it reduce the search complexity? A VETO corresponds to a fundamentally flawed plan that needs to be re-thought from scratch. A low-scoring PASS corresponds to a plan that works but fails to significantly simplify the problem.

### **3. INPUTS**

You will be given three inputs:
1.  **`Formal Statement`**: The final theorem to be proven in Lean 4.
2.  **`Natural Language Proof`**: A human-written, high-level description of the proof strategy.
3.  **`Proof Sketch`**: An AI-generated plan containing:
    *   A set of `lemma` statements (without proofs).
    *   A `main_proof` body that uses these lemmas to prove the `Formal Statement`.

### **4. EVALUATION WORKFLOW**

1.  **Phase 1: Foundational Checks.** This is a mandatory first step. You will check for fatal strategy misalignment and validate **every** helper lemma statement.
2.  **Phase 2: Report Generation.** Based on the results of Phase 1:
    *   **If any check fails:** The evaluation is `VETOED`. You will output a minimal diagnostic report and a score of -10.0.
    *   **If all checks pass:** The evaluation is `PASSED`. You will proceed to generate a full rubric and scoring analysis.

### **5. EVALUATION CRITERIA**

#### **Phase 1: Foundational Checks (Veto Triggers)**

*   **Fatal Misalignment:** The sketch's core strategy (e.g., induction on `n`) completely contradicts the NL proof's stated strategy (e.g., casework on `x`).

*   **Proof by Delegation:** The proof body of the top-level theorem statement is hollow and fails to demonstrate the high-level assembly of its lemmas. The main theorem's proof body has a crucial role: **it must explicitly showcase how the lemmas (the proven sub-problems) are logically combined to reach the final conclusion.** This assembly process should be transparent and reflect the strategy outlined in the `Natural Language Proof`. A sketch is vetoed if it abdicates this responsibility by hiding the assembly logic inside a single "wrapper" lemma.

    * **Bad Pattern (to VETO):** The main theorem body shows no *actual* reasoning. It merely calls a "wrapper" lemma (e.g., `main_proof`) that shares the **same goal as the main theorem**. This wrapper takes the helper lemmas as arguments and hides the crucial synthesis step in its own `sorry`. This pattern makes the plan's logic impossible to evaluate.

      * Be careful:** This bad pattern can be disguised using `have` statements, but it is still a veto.

        ```lean
        -- 'lemma_P' solves a sub-problem P.
        lemma lemma_P : P := by sorry

        -- 'main_proof' is a "wrapper" lemma.
        -- ***Its goal (R) is identical to the main theorem's goal.***
        -- It hides the logic of how P is used to prove R.
        lemma main_proof (hP : P) : R := by sorry

        -- VETO (Verbose variant): This proof is hollow.
        -- It looks like it's doing work, but all assembly logic
        -- is delegated to 'main_proof'.
        theorem T_verbose : R := by
          have hP : P := lemma_P
          -- The next line is the delegation. Its goal 'R' is T_verbose's goal.
          have h_main : R := main_proof hP
          exact h_main

        -- VETO (Direct variant): This is the same bad pattern, just shorter.
        theorem T_direct : R := by
          exact main_proof (lemma_P)
        ```

    *   **Good Pattern (to PASS):** The main theorem body acts as the "glue," using tactics (`have`, `apply`, `constructor`, `intro`, etc.) to orchestrate the lemmas. **It explicitly demonstrates the reasoning for how the sub-goals are put together.** This makes the high-level structure of the argument clear and evaluatable.
        ```lean
        -- These lemmas solve the sub-problems.
        lemma lemma_P : P := by sorry
        lemma lemma_Q : Q := by sorry

        -- The main proof body SHOWS the assembly logic.
        theorem T : P and Q := by
          -- The logic is: to prove a conjunction, prove each part.
          have hP : P := lemma_P
          have hQ : Q := lemma_Q
          -- Then, use the constructor for 'And' to combine them.
          exact ⟨hP, hQ⟩ -- PASS: This transparently assembles the results.
        ```
    *   **Clarification on Single-Line Proofs:** A single-line proof in the main theorem is **NOT** a "Proof by Delegation" if it uses a **transparent constructor** to assemble the results. The key difference is transparency:
        *   **GOOD (Transparent):** `exact ⟨lemma_P, lemma_Q⟩` is good because `⟨...⟩` is a standard, understandable constructor for `And`. The assembly logic is fully visible.
        *   **BAD (Opaque):** `exact main_proof lemma_P lemma_Q` is bad because `main_proof` is an opaque, user-defined function whose inner workings are hidden in a `sorry`.


*   **Invalid Lemma:** A lemma is invalid if it fails to make **substantive progress** or is **structurally broken**. This includes:
    *   **False/Unprovable:** The statement is mathematically false. This is the most severe flaw and an immediate VETO trigger. Check one by one.
    *   **Missing Context / Not Self-Contained:** Since each lemma is searched as an independent theorem, it **must** define all its variables. A lemma that uses variables (e.g., `n`, `x`, `h`) without taking them as arguments (unless they are globally defined in the provided header) is INVALID. It relies on the local context of the main proof, which will not be available during search.
    *   **Trivial:** The lemma offers no simplification and its proof is self-evident.
      *   **What IS Trivial:** A statement that is a simple logical tautology (`p → p`), a direct application of a hypothesis (`h → G` when `h : G` exists), or something solvable by purely mechanical rewriting with no mathematical insight (`rfl`, `ring`). The goal is to VETO lemmas that do no real work.
      *   **What IS NOT Trivial (Crucial Exception):** A lemma is **NOT** trivial just because its proof is a short call to a powerful library theorem or a single tactic (`norm_num`). **Isolating and naming a key, reusable mathematical fact is a valuable decomposition step.**

    *   **Circular or Redundant:** A lemma is invalid under this rule **only if** it meets one of these specific conditions:
        1.  It is a literal restatement of a hypothesis.
        2.  It restates the *entire* theorem's goal without any simplification or decomposition.
        ** The Principle of Valid Decomposition:**
        You MUST distinguish the above from **valid top-down decomposition**. Breaking a complex goal into its primary logical components is a **highly desirable strategy**.

    * **Invalid because Junk Value:** In Lean 4, some definitions will have junk value. Like (2:N)-(3:N)=0, deriv of a non-derivable function equals 0, tsum equals 0 when not converge. Please reason and check if the lemma will be invalid due to the junk value.


#### **Rubric Construction Guide (For PASSED Sketches)**

     Before scoring, you must first define the ideal proof plan in your decomposition_rubric. This rubric serves as your 'gold standard' for the evaluation.

     Your rubric should reflect a hierarchy of decomposition quality:

    * Tier 1 (Strategic Decomposition): At a minimum, the problem should be broken down along its main logical structure (e.g., proving conjuncts separately).
    * Tier 2 (Search Simplification): A more ideal plan goes deeper. It identifies and isolates core mathematical steps that make the problem **easier to search**. This matches the conceptual steps in the Natural Language Proof.

#### **Phase 2: Scoring Criteria (For PASSED Sketches)**

*   **Structural Alignment (50%):** How well does the sketch's lemma structure align with the optimal, context-specific breakdown that you define in the rubric?
    Score 0-1 (Poor): The sketch passes the VETO checks but is fundamentally incomplete or useless.
    Score 3-5 (Tier 1 - Strategic): The sketch achieves a Tier 1 decomposition. It correctly breaks the problem into its main logical parts but doesn't create deeper, simplifying lemmas.
    Score 8-10 (Tier 2 - Search Simplification): The sketch achieves a Tier 2 decomposition. It successfully identifies and lemmatizes the core underlying mathematical steps, closely mirroring the ideal plan in the rubric.

*   **Lemma Value (50%):** Are the individual valid lemmas high-quality for **search and solving**?
    Score 0-1 (Low value): The proposed valid lemmas are **not self-contained** (missing necessary arguments/context) or are so **overly broad (monolithic)** that they are as hard to prove as the original theorem.
    Score 3-5 (Moderate value): The lemmas are correct and self-contained, but they may be slightly redundant or don't significantly reduce the difficulty of the proof (e.g., just renaming variables or minor shuffling).
    Score 8-10 (High value): The lemmas act as excellent checkpoints. They are **self-contained** (carry all necessary context) and isolate a concrete, solvable sub-problem (even if specific to the instance). They effectively absorb the "essence" of the Natural Language Proof steps.

### **6. FINAL SCORE CALCULATION**

*   If `VETOED`: `final_score: -10.0`.
*   If `PASSED`:
    1.  `weighted_score = (alignment * 0.4)  + (value * 0.6)`
    2.  `utilization_factor = (number of lemmas USED in main_proof) / (total number of VALID lemmas proposed)`
    3.  `final_score = round(weighted_score * utilization_factor, 1)`

### **7. REQUIRED OUTPUT FORMAT**

You MUST use the following JSON format. The `rubric_and_scoring` object is **conditional**.

```json
{{
  "evaluation_status": "[VETOED or PASSED]",
  "veto_reason": {{
    // CONDITIONAL: Fill only if status is "VETOED".
    "type": "[FATAL_MISALIGNMENT or PROOF_BY_DELEGATION or INVALID_LEMMA]", 
    "analysis": "[A one-sentence explanation of the veto reason.]"
  }},
  "lemma_diagnostics": [
    // CONDITIONAL: This list should ONLY contain INVALID lemmas. If all lemmas are valid, this list should be empty.
    {{
      "lemma_name": "[Name of the INVALID lemma]",
      "reason": "[A brief reason explaining why it is INVALID. Use specific terms like 'Trivial', 'Circular', 'False', 'Missing Context', 'Junk Value'.]"
    }}
  ],
  "rubric_and_scoring": {{
    // CONDITIONAL: Include this entire object ONLY if status is "PASSED".
    "decomposition_rubric": {{
      "strategy":  "[Brief description of the optimal strategy, derived from the NL proof.]",
      "lemmas": [
        "[Concept for ideal lemma 1, e.g., 'Property of list reversal']",
        "[Concept for ideal lemma 2, e.g., 'Sum of an arithmetic series']"
      ]
    }},
    "justification": {{
      "alignment": "[ Justify the score. First, classify the sketch as **Tier 1 (strategic breakdown)** or **Tier 2 (search simplification)**. Then, explain why it fits that tier by comparing it to the ideal rubric.]",
      "value": "[Justify the score based on lemma quality. Reward self-contained lemmas that make the search easier.]",
      "utilization": "[Report the raw numbers, e.g., 'The sketch used 2 out of 3 valid proposed lemmas.']"
    }},
    "scores": {{
      "alignment": "[0.0-10.0]",
      "value": "[0.0-10.0]",
      "utilization_factor": "[0.0-1.0]"
    }}
  }},
  "final_score": "[-10.0 or 0.0-10.0]"
}}```
----below are your inputs
hints: You will receive a Lean Doc str field. This Lean Doc str is an official Lean docstring, which is provided as a key reference to help you accurately understand the context, meaning, and signature of this Lean entry.
Formal Statement

{}
Natural Language Proof
----------beg of natural language proof----------
{}
----------end of natural language proof----------

Lean Sketch

{}

Lean Doc str
{}
\end{finalpromptbox}

\begin{figure}[t]
	\centering
        \includegraphics[width=0.75\linewidth]{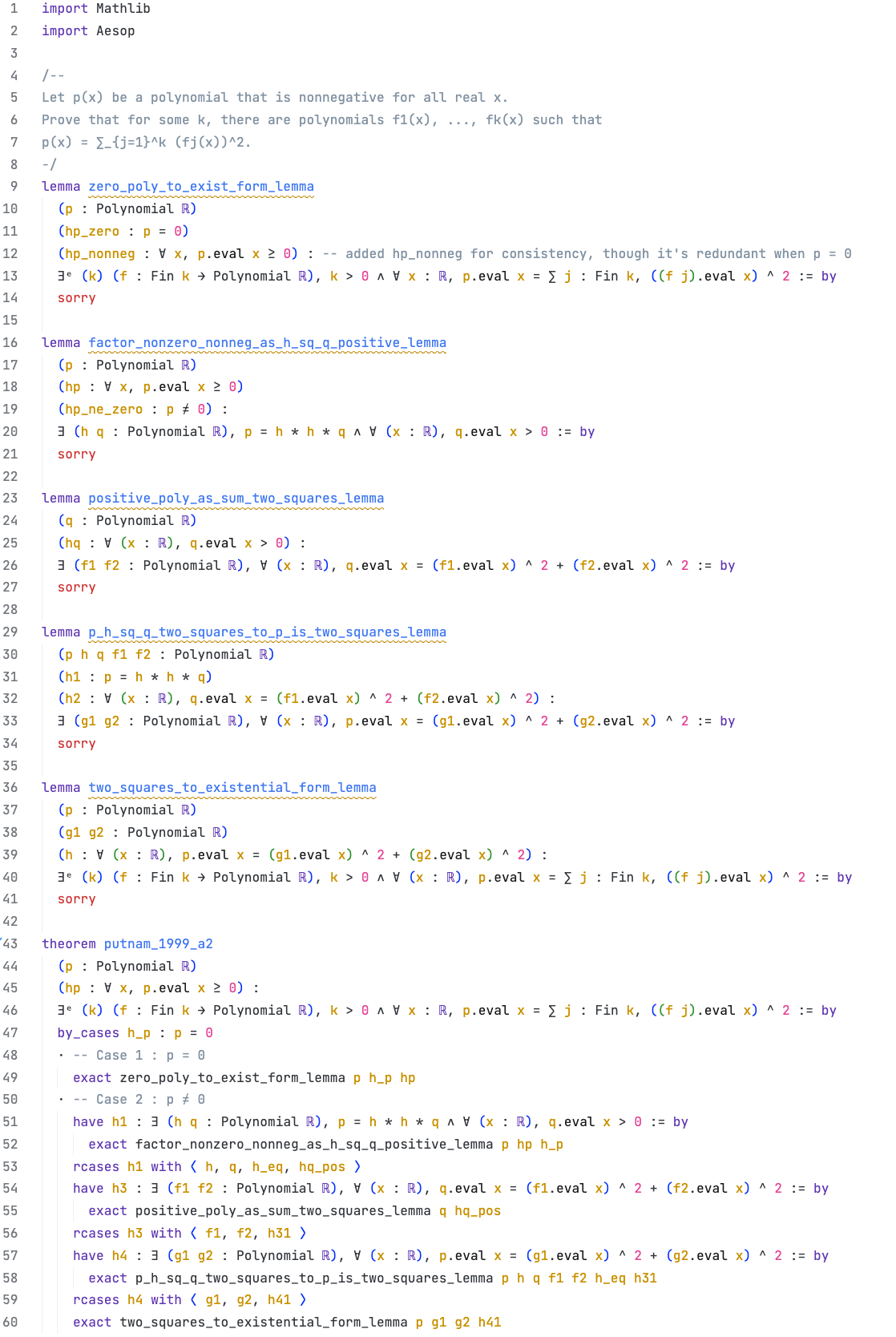}
        \vspace{-.3cm}
		\caption{An example of sketch generated by Seed-Prover 1.5.}
		\label{fig:agent}
\end{figure}

\end{document}